\pdfoutput=1
\documentclass[11pt]{article}
\usepackage[]{acl}
\usepackage{microtype}

\usepackage{times}
\usepackage{latexsym}
\usepackage{lipsum}
\usepackage{booktabs}
\usepackage[labelfont=bf]{caption}
\usepackage{multirow}
\usepackage{graphicx}
\usepackage{amssymb}
\usepackage{amsmath}
\usepackage{enumerate}
\usepackage{soul}
\usepackage[T1]{fontenc}

\usepackage[utf8]{inputenc}
\usepackage{inconsolata}

\usepackage{graphicx} % DO NOT CHANGE THIS
\usepackage{natbib}  % DO NOT CHANGE THIS AND DO NOT ADD ANY OPTIONS TO IT
\usepackage{caption} % DO NOT CHANGE THIS AND DO NOT ADD ANY OPTIONS TO IT
\usepackage{multirow}
\usepackage{algorithm}
\usepackage{algpseudocode}
\usepackage{amsmath}
\usepackage{booktabs}
\usepackage{amsfonts}
\usepackage{blkarray, bigstrut}
\parskip=\medskipamount
\usepackage{amssymb}% http://ctan.org/pkg/amssymb
\usepackage{pifont}% http://ctan.org/pkg/pifont

\newcommand{\fl}{{\tt fluency}}
\newcommand{\co}{{\tt coherence}}
\newcommand{\rel}{{\tt relevance}}
\newcommand{\fa}{{\tt faithfulness}}
\newcommand{\asp}{{\tt aspect coverage}}
\newcommand{\sent}{{\tt sentiment consistency}}
\newcommand{\spec}{{\tt specificity}}

\newcommand{\gptfourstandard}{{\tt GPT-4}}

\usepackage[labelfont=bf]{caption}
\usepackage{multirow}

\usepackage{newfloat}
\usepackage{listings}

\usepackage{xcolor}
\usepackage[capitalise,noabbrev]{cleveref}
\usepackage{hyperref}

\newcommand\blfootnote[1]{%
  \begingroup
  \renewcommand\thefootnote{}\footnote{#1}%
  \addtocounter{footnote}{-1}%
  \endgroup
}

\title{LLMs as Architects and Critics for Multi-Source Opinion Summarization}

\author{Anuj Attri$^\diamondsuit$$^\clubsuit$,
Arnav Attri$^\diamondsuit$$^\clubsuit$, Pushpak Bhattacharyya$^\clubsuit$ 
\\ 
\textbf{Suman Banerjee$^\mathcal{F}$, Amey Patil$^\mathcal{F}$, Muthusamy Chelliah$^\mathcal{F}$, Nikesh Garera$^\mathcal{F}$}\\
\textbf{}
        $^\clubsuit$Computer Science and Engineering, IIT Bombay, India, 
        $^\mathcal{F}$Flipkart, India \\
        \texttt{\{arnavcs, ianuj,
        pb\}@cse.iitb.ac.in}
        }

\begin{document}
\maketitle
\blfootnote{$^\diamondsuit$ Equal contribution}
\begin{abstract}

\textls[-60]{\textsc{Multi-source Opinion Summarization}} \textls[-60]{\textsc{(M-OS)}} extends beyond traditional opinion summarization by incorporating additional sources of product metadata such as \texttt{descriptions, key features, specifications, and ratings, alongside reviews}. This integration results in comprehensive summaries that capture both subjective opinions and objective product attributes essential for informed decision-making. While Large Language Models (LLMs) have shown significant success in various Natural Language Processing (NLP) tasks, their potential in \textls[-60]{\textsc{M-OS}} remains largely unexplored. Additionally, the lack of evaluation datasets for this task has impeded further advancements. To bridge this gap, we introduce \textls[-60]{\textsc{M-OS-EVAL}}, a benchmark dataset for evaluating multi-source opinion summaries across $7$ key dimensions: \textls[-60]{\textsc{\fl}}, \textls[-60]{\textsc{\co}}, \textls[-30]{\textsc{\rel}}, \textls[-60]{\textsc{\fa}}, \textls[-30]{\textsc{\asp}}, \textls[-30]{\textsc{\sent}}, \textls[-30]{\textsc{\spec}}. Our results demonstrate that \textls[-60]{\textsc{M-OS}} significantly enhances user engagement, as evidenced by a user study in which, on average, $\textbf{87\%}$ of participants preferred \textls[-60]{\textsc{M-OS}} over opinion summaries. Our experiments demonstrate that factually enriched summaries enhance user engagement. Notably, \textsc{M-OS-PROMPTS} exhibit stronger alignment with human judgment, achieving an average Spearman correlation of $\boldsymbol{\rho = 0.74}$, which surpasses the performance of previous methodologies.

\end{abstract}

\section{Introduction}

\begin{table}[h!]
\centering
\resizebox{1\columnwidth}{!}{
\begingroup
\renewcommand{\arraystretch}{1.2}
\Large
\begin{tabular}{lcccccccc}
    \toprule
        \multicolumn{9}{c}{\textit{Opinion Summary}} \\
    \midrule
        \multicolumn{9}{p{15cm}}{
        %The racquet is highly praised for its lightweight design and powerful performance, particularly in smashes, making it ideal for both offensive and defensive play. Many users recommend it for intermediate to advanced players, especially those on a budget. The head-heavy design enhances power without compromising ease of movement, and its construction quality, including military-grade carbon, is appreciated. Some users note minor drawbacks, such as a smaller grip size, which can easily be adjusted with external grips. While M-OSt feedback is overwhelmingly positive, one customer reported receiving a broken racquet with no applicable replacement policy. Overall, the racquet is considered a great value for its price, offering excellent performance and durability, with its lightweight nature being a key highlight.
        The racquet is highly praised for its lightweight design and powerful performance, particularly in smashes, making it ideal for both offensive and defensive play. Many users recommend it for intermediate to advanced players, especially those on a budget. The head-heavy design boosts power while ensuring easy movement, with durable military-grade carbon construction. Some users note minor drawbacks, such as a smaller grip size, which can easily be adjusted with external grips. However, one customer reported receiving a damaged racquet with no replacement policy. The racquet offers excellent performance, durability, and a lightweight design.
        } \\
    \midrule
        \multicolumn{9}{c}{\textit{Multi-Source Opinion Summary}} \\
    \midrule
        \multicolumn{9}{p{15cm}}{
        The Li-Ning G-Force Racquet is designed for advanced players prioritizing speed and precision. Crafted from \textbf{Japanese Ultra-Carbon Graphite}, it ensures exceptional strength and rigidity, preventing frame deformation. The \textbf{UHB Shaft} designed using player data, optimizes smash performance, while the \textbf{oval 53.5 sq/in head size and G2 grip (9.52 cm)} provide control and maneuverability. The \textbf{Aero Tec Beam System} reduces air resistance while preserving frame integrity, and the \textbf{Dynamic Optimum Frame design} boosts bounce strength for powerful smashes. \textbf{Weighing only 79g} with a head-heavy balance, it excels in both offensive and defensive play. At \textbf{26.7 inches tall} with a \textbf{7 mm beam width}, it delivers top-tier performance. With an \textbf{average rating of 4.2}, users praise its durability, though some recommend external grips. Despite a rare issue with a broken racquet, it remains a top choice for players seeking precision and power.} \\
    \bottomrule
\end{tabular}
\endgroup
}

\caption{Opinion Summary vs. Multi-Source Opinion Summary (M-OS) by \texttt{GPT-4o} for a Badminton Racquet. While the opinion summary from reviews alone provides subjective feedback, the M-OS integrates product metadata with reviews to deliver a comprehensive overview. This eliminates manual metadata parsing while maintaining balanced product coverage. \textbf{Boldface} indicates technical specifications absent in the opinion summary.}

\label{tab:summary_comparison}
\end{table}

%In the dynamic landscape of e-commerce, customer reviews play a pivotal role in influencing purchasing decisions. However, the sheer volume of reviews per product makes it challenging for users to process all the available information efficiently. Opinion summarization addresses this issue by distilling the content of reviews into concise summaries \cite{wang2016neural,chu2019meansum,braz2020_int}. Despite its utility, traditional approaches to opinion summarization focus primarily on user reviews, often overlooking other valuable sources of product information.

Customer reviews, while crucial for e-commerce decisions, present challenges due to their overwhelming volume. Traditional opinion summarization approaches \cite{wang2016neural,chu2019meansum,braz2020_int} generate concise summaries but rely solely on user reviews, missing valuable product information from other sources.

We introduce Multi-Source Opinion Summarization (M-OS), which integrates reviews with product descriptions, specifications, and ratings to create comprehensive summaries. M-OS combines subjective user experiences with objective product attributes to facilitate informed decision-making.
As demonstrated in Table \ref{tab:summary_comparison}, M-OS enriches summaries by incorporating technical specifications and product descriptions, enabling precise product comparisons - a key advantage over review-only approaches that often lack detailed attributes.
\textsc{M-OS} addresses decision fatigue and information overload by synthesizing diverse product data to deliver comprehensive, relevant summaries. This streamlined approach reduces cognitive load and enhances user satisfaction by providing actionable insights without requiring manual metadata parsing.

%\begin{figure}[t]
%    \centering
 %   \includegraphics[width=1\columnwidth]{first.png}
  %  \caption{Traditional Opinion Summary and M-OS generated by {\tt GPT-4o}} 
%    \label{fig:first}
%\end{figure}
LLMs have emerged as effective reference-free evaluators for NLG tasks \citep{fu, chiang-lee-2023-large, closer, wanggpt, kocmi-federmann-2023-large}, addressing the limitations of traditional metrics like ROUGE \citep{ro} and BERTSCORE \cite{zhang-etal-2019-pretraining} which correlate poorly with human judgments \cite{shen2023opinsummeval}. Given the high costs of reference datasets and the inadequacy of conventional metrics for multi-source opinion summaries, LLM-based evaluation offers a scalable solution. We present \textsc{M-OS-EVAL}, a reference-free evaluation dataset for multi-source opinion summarization that assesses summaries across $7$ dimensions through two frameworks: \textsc{Omni-Prompt}, a dimension-independent prompt, and \textsc{Spectra-Prompts}, a dimension-dependent prompt set.

To address this need, we present \textsc{M-OS-EVAL}, a benchmark dataset for evaluating M-OS across $7$ key dimensions: \textls[-60]{\textsc{\fl}}, \textls[-60]{\textsc{\co}}, \textls[-30]{\textsc{\rel}}, \textls[-60]{\textsc{\fa}}, \textls[-30]{\textsc{\asp}}, \textls[-30]{\textsc{\sent}}, \textls[-30]{\textsc{\spec}}. We propose two novel evaluation frameworks: \textsc{Omni-Prompt}, which enables metric-independent assessment, and \textsc{Spectra-Prompts}, which facilitates metric-dependent evaluation across all $7$ dimensions. Our work represents the first prompt-based evaluation method for M-OS, incorporating both closed-source and open-source models to advance LLM-based evaluation in this domain.

\begin{itemize}
    \item \textbf{\textls[-30]{\textsc{M-OS:}}} \textls[-30]{\textsc{Multi-Source Opinion Summarization (or Summary)}}.
    \item \textbf{\textls[-50]{\textsc{M-OS-GEN:}}} \textls[-50]{\textsc{Multi-Source Opinion Summary Generation}}.
    \item \textbf{\textls[-50]{\textsc{M-OS-EVAL:}}} \textls[-50]{\textsc{Multi-Source Opinion Summary Evaluation}}.
\end{itemize}

Our contributions are:

\begin{enumerate}
\item \textbf{M-OS:} We advance multi-source opinion summarization by using LLMs to generate comprehensive summaries that integrate product metadata (title, description, features, specifications, rating) with customer reviews. Unlike \textsc{\textbf{MEDOS}} \citep{siledar2024b}, M-OS dynamically synthesizes unified summaries that present essential product information upfront, eliminating the need for users to parse metadata separately. Our user study shows that 87\% of participants on average found multi-source summaries more comprehensive than opinion summaries (Section \ref{study})

\item \textbf{M-OS-DATA:} A novel dataset of $25,000$ unique products across diverse categories, each with comprehensive metadata, enabling robust training and evaluation of multi-source opinion summarization models (Section \ref{M-OS-data}).
\item \textbf{M-OS-EVAL}: A comprehensive evaluation benchmark comprising $4,900$ summary annotations across $7$ key dimensions: \textls[-60]{\textsc{\fl}}, \textls[-60]{\textsc{\co}}, \textls[-30]{\textsc{\rel}}, \textls[-60]{\textsc{\fa}}, \textls[-30]{\textsc{\asp}}, \textls[-30]{\textsc{\sent}}, \textls[-30]{\textsc{\spec}}, for thorough assessment of multi-source opinion summaries (Section \ref{M-OS-eval}).
\item We introduce \textbf{\textsc{Omni-Prompt}} for metric-independent assessment and \textbf{\textsc{Spectra-Prompts}} for metric-dependent evaluation across  aforementioned $7$ dimensions. This represents the first prompt-based method for assessing multi-source opinion summarization and evaluating various open-source LLMs in this domain (Section \ref{method}).

\item Benchmarking of $14$ recent LLMs (closed and open-source) on the aforementioned $7$ dimensions for the task of multi-source opinion summarization, which to the best of our knowledge is first of
its kind (Table \ref{tab:M-OS_human}, Section \ref{results_analysis}).

\item We compare four open-source LLMs against a closed-source (\textsc{GPT-4o}) LLM for automatic M-OS evaluation across $7$ dimensions. Our analysis reveals M-OS-PROMPTS as an effective alternative, achieving strong alignment with human assessment (average Spearman correlation: $0.74$) (Table \ref{tab:_M-OS_evaluator_inp}, Section \ref{results_analysis}).

\end{enumerate}

\section{Related Work}
\begin{figure*}[htp]
    \centering
    \includegraphics[width=2\columnwidth]{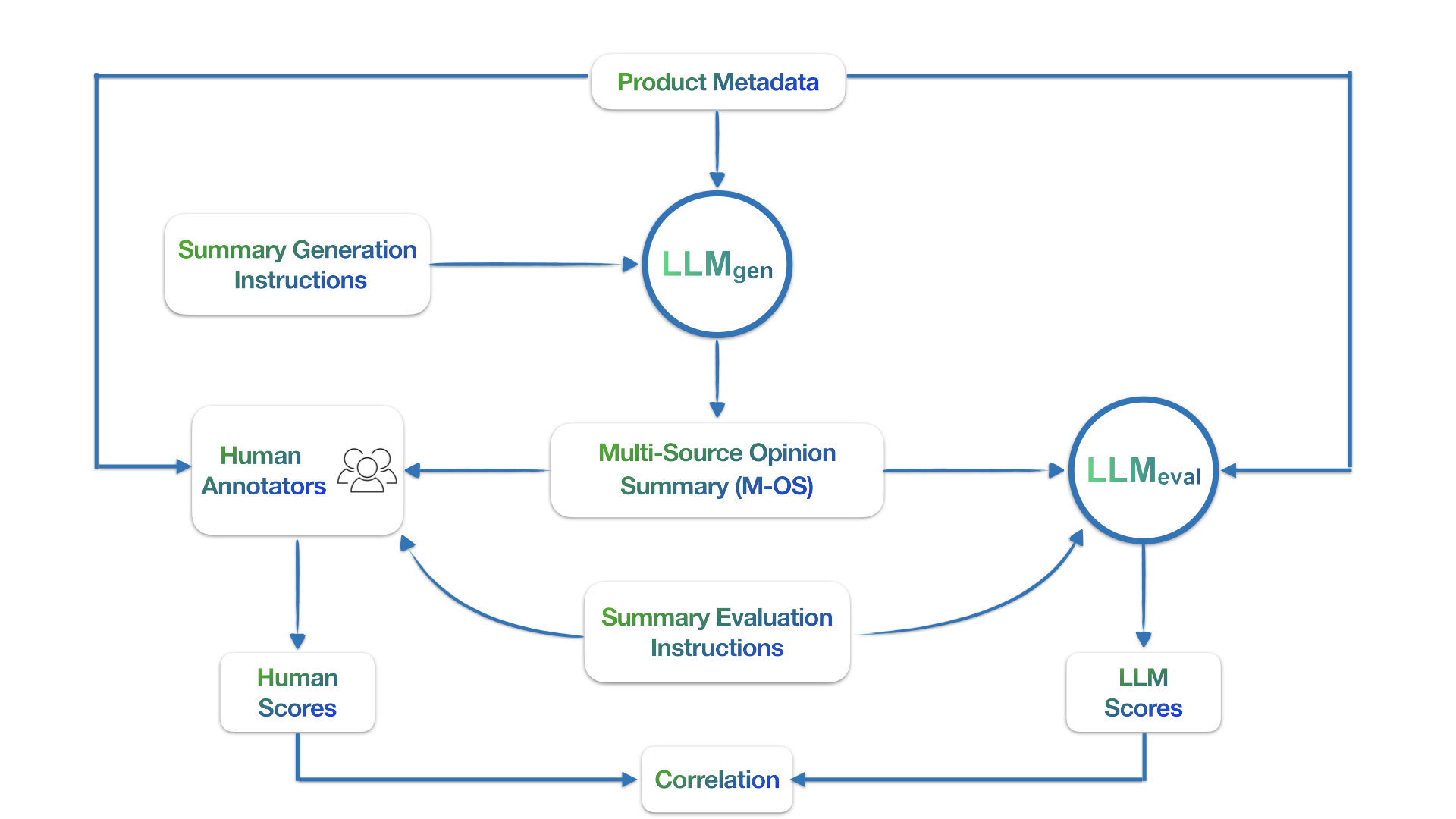}
    \caption{ Pipeline of our experiments involving generation, evaluation, and correlation analysis. \textbf{LLM\(_{\text{gen}}\)} \textit{(LLM used as a generator)}, which generates M-OS using product metadata, \textit{(comprising product title, description, key features, specifications, average ratings, and customer reviews)} guided by the Summary Generation Instructions \textsc{(M-OS-GEN-PROMPT)}. \textbf{LLM\(_{\text{eval}}\)} \textit{(LLM used as an evaluator)}, which evaluates the summaries based on Summary Evaluation Instructions \textsc{(M-OS-EVAL-PROMPT)}.
    }
    \label{fig:big}
\end{figure*}

Opinion summarization has evolved from extractive methods \cite{erkan2004lexrank,kim2011comprehensive} to neural-based approaches \cite{braz2020_int, amplayo2020unsupervised}, with various specialized directions emerging. For aspect-specific summarization, \citet{angelidis2021extractive} employed VQ-VAE \citep{vandenOord2017}, while \citet{amplayo2021} introduced abstractive approaches using MIL. Self-supervised methods were advanced by \citet{brazinskas2020} using pseudo-summary pairs, enhanced by \citet{amplayo2020} with noisy variations and \citet{elsahar2021} with TF-IDF similarity-based selection. Large-scale processing was addressed by \citet{bhaskar2023} using GPT-3.5 \citep{openai2023} prompting, \citet{jiang-etal-2023-large} with sentiment-aware sampling, and \citet{muddu2024distillingopinionsscaleincremental} through \textsc{XL-OPSUMM}. Multi-source approaches emerged with \cite{Zhao_Chaturvedi_2020} utilizing product descriptions, \cite{Li_Yuan_Xu_Wu_He_Zhou_2020} developing supervised multimodal methods, and \citet{siledar2024b} introducing a structured approach with reviews, descriptions, and Q\&A pairs. While these methods advanced the field, they typically overlook comprehensive product metadata. Our work differs by leveraging LLMs' extended context lengths to incorporate complete product specifications, generating comprehensive summaries that eliminate the need for manual navigation through product information. We extend beyond \citet{siledar2024b} by including detailed specifications and descriptions, providing users with complete product insights in a single, unified summary.

\textbf{LLM-based Evaluators} Traditional metrics like ROUGE \citep{ro}, BLEU \citep{bleu} and BERTSCORE \cite{zhang-etal-2019-pretraining} correlate poorly with human judgments \cite{shen2023opinsummeval}. LLM-based evaluation provides a cost-effective solution for large-scale reference-based datasets, including Chain of Thought approaches \citep{liuug, wei2023}, reference-free evaluation \citep{chianggg}, and other methods \citep{fu, chiang-lee-2023-large, closer, wanggpt, kocmi-federmann-2023-large}. \citet{op} proposed two prompt strategies for opinion summarization. We leverage LLMs as evaluators for reference-free \textsc{M-OS} assessment.

\section{Methodology}\label{method}
Our methodology is centered around the development of \textsc{M-OS-PROMPTS}, which facilitate both the \textit{generation} and \textit{evaluation} of M-OS.

\subsection{M-OS-GEN-PROMPT (Summary Generation Prompt)}
The \textsc{M-OS-GEN-PROMPT} guides LLMs to create summaries by synthesizing information from various product attributes, including the product title, description, key features, specifications, customer reviews, and average ratings. By integrating these diverse data sources, the prompt ensures that summaries provide a comprehensive perspective, balancing subjective customer opinions with objective product details. For M-OS-GEN-PROMPT design principles, refer to Appendix ~\ref{gen_design_principle}.

\subsection{M-OS-EVAL-PROMPTS (Summary Evaluation Prompts)}
The \textsc{M-OS-EVAL-PROMPTS} guide evaluation of \textsc{M-OS}, structured to assess $7$ dimensions: \textls[-60]{\textsc{\fl}}, \textls[-60]{\textsc{\co}}, \textls[-30]{\textsc{\rel}}, \textls[-60]{\textsc{\fa}}, \textls[-30]{\textsc{\asp}}, \textls[-30]{\textsc{\sent}}, \textls[-30]{\textsc{\spec}} and \texttt{specificity} (Appendix \ref{metric_definition} for metric definition). Each prompt has four core components for thorough evaluation across open-source and closed-source LLMs:

\noindent(\textit{1}) {\tt \textbf{System Message}}: Defines the LLM's role as a specialized evaluator, providing clear context for the evaluation task.

\noindent(\textit{2}) {\tt \textbf{Task Description}}: Outlines the specific evaluation task, which involves assessing a multi-source opinion summary against product metadata (including title, description, key features, specifications, reviews, and average rating).

\noindent(\textit{3}) {\tt \textbf{Evaluation Criteria}}: Defines the criteria for the task. For multi-source opinion summary evaluation, the LLM assigns a score $(1-5)$ based on how well the summary adheres to a specific metric or dimension.

\noindent(\textit{4}) {\tt \textbf{Evaluation Step}}: Provides the LLM with a detailed, step-by-step guide to complete the evaluation, ensuring consistency and thoroughness.

For M-OS-EVAL-PROMPTS design principles, refer to Appendix \ref{eval_design_principle}.

\noindent \paragraph{M-OS-EVAL PROMPT Approaches} 
%We extend the prompt strategies introduced by \citet{op}, we develop two approaches specifically engineered for M-OS evaluation:
Building on \citet{op}, we develop two M-OS evaluation approaches.For graphical visualization, refer to Appendix \ref{prompt_view}.

%\textbf{\textsc{Omni-Prompt}} (universal prompt for comprehensive cross-dimensional evaluation) represents our metric-independent evaluation approach. The key innovation lies in its modular architecture: while maintaining a consistent framework of Task Description, Evaluation Criteria, and Evaluation Steps, it introduces a flexible 'Metric' component that can be dynamically modified. This design enables universal applicability—the same prompt structure can evaluate any dimension by simply redefining the 'Metric' component, while the surrounding framework ensures methodological consistency across all evaluations.
\textbf{\textsc{Omni-Prompt}} (universal prompt for comprehensive cross-dimensional evaluation) represents our metric-independent evaluation approach with a modular architecture. While maintaining a consistent framework of Task Description, Evaluation Criteria, and Evaluation Steps, it introduces a flexible 'Metric' component for dynamic modification. This enables universal applicability—the same structure evaluates any dimension by redefining the 'Metric' component while ensuring methodological consistency.

%\textbf{\textsc{Spectra-Prompts}} (prompts for nuanced, criterion-specific analysis) comprise our suite of dimension-specific evaluation prompts. Each prompt is precisely engineered for one of the $7$ evaluation dimensions, incorporating specialized criteria and assessment guidelines. While this approach demands deeper expertise in both the evaluation domain and prompt engineering, it offers unparalleled precision in dimension-specific assessment. %Unlike the metric-independent approach, these prompts cannot be repurposed across dimensions, but they offer more targeted and nuanced evaluation for each specific criterion.
%This specialization makes \textsc{Spectra-Prompts} particularly effective for detailed analysis of complex multi-source summaries where different dimensions require distinct evaluation approaches. These prompts operate independently, with each crafted to capture the unique nuances of its target dimension, though this specialization means they cannot be repurposed across dimensions.
\textbf{\textsc{Spectra-Prompts}} (prompts for nuanced, criterion-specific analysis) comprises dimension-specific evaluation prompts, each engineered for one of the $7$ evaluation dimensions with specialized criteria and assessment guidelines guidelines. While requiring deeper expertise in evaluation and prompt engineering, it offers unparalleled precision in dimension-specific assessment. These prompts operate independently to capture unique dimensional nuances but cannot be repurposed across dimensions.

Our evaluation compares these approaches with \textsc{Op-I-Prompt} and \textsc{Op-Prompts}, the current state-of-the-art in prompt-based summary evaluation.

\subsection{Scoring Function}\label{scoring_func}
%\citet{liuug} identified a limitation in relying on LLMs to output a single integer score, suggesting that a weighted average of scores offers a more robust evaluation. In this approach, the weights are determined by the probabilities assigned to each score by the LLM. Formally, given a set of scores $\{s_1, \dots, s_j\}$, the probability of each score $p(s_{k})$ is estimated by the LLM, and the final score $o$ is computed as:
%\begin{align} 
%    o = \sum_{k=1}^{j} p(s_{k})\times s_{k} 
%\end{align}
%Here, $p(s_{k})$  represents the probability of score $s_{k}$ for a given input, as estimated by the LLM. To derive these probabilities, we sample $n$ outputs from the LLM, where $n$ is large enough ($\sim 100$) to provide a reliable estimate. The scoring function, in this case, simplifies to taking the mean over the $n$ outputs, ensuring a more accurate and probabilistically grounded final score.

%By adopting this method, we account for the uncertainty in LLM outputs, allowing the scoring process to better capture the nuances of LLM evaluations and avoid the limitations of single-point estimates.
\citet{liuug} proposed a weighted average approach to address discrete LLM scoring limitations. The final score is computed as:
\begin{align} 
    o = \sum_{k=1}^{j} p(s_{k})\times s_{k} 
\end{align}
where $s_{k}$ are possible scores and $p(s_{k})$ their LLM-determined probabilities. $p(s_{k})$ is estimated by sampling $n$ outputs $({\tt n} \approx 100)$  per input, effectively reducing scoring to a mean calculation. This method aims to enhance scoring nuance and reliability by addressing the inherent uncertainty in LLM outputs. By incorporating this approach, the scoring process captures the subtleties of LLM evaluations more effectively, mitigating the limitations of single-point estimates.

\subsection{Evaluation Approach}\label{appendix_correlation}

For each product $p_{i}$ in dataset $\mathcal{D}$, $i \in \{1,...,\mathcal{Q}\}$, we have $\mathcal{N}$ \textsc{M-OS} from different models. Let $s_{ij}$ denote the $j^{th}$ \textsc{M-OS} for product $p_{i}$, $\mathcal{M}_m$ denote the $m^{th}$ evaluation metric and $\mathcal{K}$ denote the correlation measure. \citet{bhandari-etal-2020-evaluating} defines the summary-level correlation as:
\begin{align}
    \mathcal{R}(a,b) = \frac{1}{\mathcal{Q}} \sum_{i} &\mathcal{K}([\mathcal{M}_{a}(s_{i1}),...,\mathcal{M}_{a}(s_{i\mathcal{N}})], \nonumber \\
    &[\mathcal{M}_{b}(s_{i1}),...,\mathcal{M}_{b}(s_{i\mathcal{N}})])
\end{align}
Where: $\mathcal{Q}$ is the total number of products %(50 in our experiments)
$s_{ij}$ is the \textsc{M-OS} generated for product $p_{i}$ by model $j$
$\mathcal{M}_a$ and $\mathcal{M}_b$ are two different evaluation metrics.

\section{Dataset} \label{dataset} 
We describe the datasets used in our study as:
\subsection{M-OS-DATA (Product Metadata Dataset)} \label{M-OS-data} 
%\textsc{M-OS-DATA} is a novel proprietary dataset specifically designed to support multi-source opinion summarization tasks. The dataset consists of products from a wide range of domains, including electronics, home \& kitchen, sports, clothing, shoes \& jewelry, among others. Each product entry comprises a rich set of metadata attributes, including but not limited to: title, description, key features, specifications, reviews, and average rating, reflecting a balance of objective details and subjective feedback. Detailed statistics of \textsc{M-OS-DATA} can be found in Table \ref{tab:dataset_stats}.
\textsc{M-OS-DATA} is a new proprietary dataset comprising products across diverse domains (electronics, home \& kitchen, sports, clothing, shoes \& jewelry, among others.). Each entry contains comprehensive metadata: title, description, features, specifications, reviews, and average rating. Statistics are presented in Table \ref{tab:dataset_stats}. The dataset was developed through a formal collaboration between our University lab and a major e-commerce company (details withheld for review anonymity). The data collection process was rigorous, senior data scientists curated the dataset using automated quality filters and manual verification to ensure data authenticity, completeness, and real-world applicability. Each product entry underwent multiple validation checks for correctness of specifications, coherence of reviews, and overall data quality. This meticulous curation process ensures the dataset's reliability for M-OS task.
\begin{table}[htp]
    \centering
    \resizebox{1\columnwidth}{!}{%
    \begin{tabular}{lcc}
         \toprule
        \textbf{Statistic} & \textbf{Value} \\
        \midrule
        \# of unique queries & 7752 \\
        Total \# of products & 23256 \\
        Average \# of reviews per product & 10 \\
        Average length of specifications per product (words) & 242.6 \\
        Average length of reviews per product (words) & 17.99 \\
        Average length of description per product (words) & 105.79 \\
        Average length of key features per product (words) & 24.64 \\
        \bottomrule
    \end{tabular}%
    }
    \caption{\textsc{M-OS-DATA} dataset statistics.}
    \label{tab:dataset_stats}
\end{table}%

%\subsubsection{Dataset Composition} Each product entry in the M-OS-DATA is associated with a variety of metadata attributes that collectively provide a holistic view of the product. These attributes include: 
%\begin{itemize} 
%\item \textbf{Product Titles}: Descriptive titles that provide an overview of the product's primary function. 
%\item \textbf{Descriptions}: Detailed textual descriptions that often elaborate on the product’s unique selling points, materials, and usage scenarios.
%\item \textbf{Key Features}: A list of highlighted features that distinguish the product from competitors, including important technical specifications for electronics or material information for clothing. 
%\item \textbf{Specifications}: Detailed Objective data points, such as dimensions, weight, and technical properties, which provide factual information about the product's physical or functional attributes. 
%\item \textbf{Customer Reviews}: User-generated content that offers subjective opinions based on actual usage, capturing a wide range of customer experiences and satisfaction levels.
%\item \textbf{Average Rating}: Aggregated numerical rating reflecting the general customer sentiment towards the product, though we note that individual reviews offer a more granular insight.
%\end{itemize}

\subsection{M-OS-EVAL (Evaluation Benchmark Dataset)} \label{M-OS-eval} 
We developed M-OS-EVAL to evaluate summaries across $7$ dimensions defined in Appendix \ref{metric_definition}. The dataset includes $14$ model-generated summaries per product for $50$ products from the M-OS-DATA test set, resulting in $14,700$ total ratings ($3$ raters × $50$ products × $14$ summaries × $7$ dimensions). Three experienced raters (Master's, Pre-Doctoral, Doctoral) evaluated each summary on a $5$-point Likert scale.

Expert raters were chosen over crowd workers based on \cite{gillick} and \cite{fabbri2021}, who demonstrated that expert annotations are superior for mitigating quality concerns. Like \cite{fabbri2021}, we conducted two rounds of evaluation; in Round II, ratings differing by $2$ or more points were re-evaluated through discussion until discrepancies were reduced to $1$ point or less.

Our raters, male students aged $24-32$, had relevant publications or active research in opinion summarization or are working in the opinion summarization domain. They received appropriate stipends. To avoid bias, model identities were undisclosed.

\subsection{Annotation Analysis} \label{anno_analysis} 
\begin{table}[t]
\centering
    \resizebox{1\columnwidth}{!}{%
    \begin{tabular}{lcc}
         \toprule
        & \textbf{Round-I} $\uparrow$ & \textbf{Round-II} $\uparrow$\\
        \midrule
        \fl & $0.73$ & $0.88$ \\
        \co & $0.67$ & $0.82$ \\
        \rel & $0.69$ & $0.85$ \\
        \fa & $0.79$ & $0.91$ \\
        \asp & $0.77$ & $0.89$ \\
        \sent & $0.66$ & $0.86$ \\
        \spec & $0.61$ & $0.84$ \\
        \midrule
        \textbf{\textsc{AVG}} & $0.70$ & $0.86$ \\
        \bottomrule
    \end{tabular}
    }
    \caption{\textbf{Inter-rater agreement scores} for Round-I and Round-II across $7$ dimensions. An improvement in agreement scores is observed in Round-II.}
    \label{tab:inter_rater_agreement}
\end{table}

%We assessed inter-rater agreement using Krippendorff's alpha coefficient ($\alpha$) \citep{Krippendorff2011ComputingKA}. In Round-I, ($\alpha$) was $0.70$ indicating moderate agreement ($0.61 \le \alpha \le 0.80$), while in Round-II, it increased to $0.86$, reflecting substantial agreement ($0.81 \le \alpha \le 1.00$). Dimension-wise agreement scores for both rounds are presented in Table \ref{tab:inter_rater_agreement}.

%\textls[-60]{\textsc{\fa}} and \textls[-30]{\textsc{\asp}} achieved the highest agreement across both rounds. \textls[-60]{\textsc{\fa}} benefited from the clear verifiability of factual content against source product metadata, while \textls[-30]{\textsc{\asp}} showed strong agreement due to annotators' ability to cross-examine the reviews and verify major aspects being discussed.

%\textls[-60]{\textsc{\co}} and \textls[-30]{\textsc{\spec}} had relatively lower agreement in Round-I, likely due to the inherent subjectivity in assessing narrative flow and the detailed nature of product-specific information. \textls[-30]{\textsc{\rel}} showed moderate agreement initially, reflecting the inherent subjectivity in determining what constitutes important information across diverse product metadata. However, it demonstrated substantial improvement in Round-II, attributed to the comprehensive annotation guidelines that helped standardize the assessment of information importance.

We measured inter-rater agreement using Krippendorff's alpha coefficient ($\alpha$) \citep{Krippendorff2011ComputingKA}. Round-I achieved $\alpha = 0.70$ (moderate: $0.61 \leq \alpha \leq 0.80$), while Round-II reached $\alpha = 0.86$ (substantial: $0.81 \leq \alpha \leq 1.00$). Table \ref{tab:inter_rater_agreement} presents dimension-wise scores.
\textls[-60]{\textsc{\fa}} and \textls[-30]{\textsc{\asp}} showed highest agreement across rounds. \textls[-60]{\textsc{\fa}}'s high agreement stemmed from verifiable product metadata, while \textls[-30]{\textsc{\asp}}'s strength came from cross-examination of reviews and major aspects.
\textls[-60]{\textsc{\co}} and \textls[-30]{\textsc{\spec}} had lower Round-I agreement due to subjective narrative assessment and detailed product information. \textls[-30]{\textsc{\rel}} improved from moderate to substantial agreement in Round-II through comprehensive guidelines that standardized importance assessment across product metadata.\textls[-30]{\textsc{\sent}} maintained steady agreement across rounds, reflecting effective criteria for sentiment alignment between summaries and reviews. \textls[-60]{\textsc{\fl}} showed stable agreement, indicating clear consensus on linguistic assessment. The improved overall agreement from Round-I to Round-II validates our evaluation framework's robustness across dimensions.

\section{Experiments} \label{exp}
Our evaluation comprises two components:

\subsection{M-OS-GEN (Summary generation)}\label{M-OS_gen}
Below is the description of Model Selection and Categorization.

\textbf{Task-specific models:} While models like \texttt{\textbf{MeanSum}} \cite{chu2019meansumn}, \texttt{\textbf{CopyCat}} \cite{brazinskas-etal-2020-unsupervised}, and \texttt{\textbf{OpinionDigest}} \cite{opiniondigest} perform well for standard opinion summarization with limited reviews, they struggle with M-OS tasks. Trained on smaller, review-only datasets, these models lack the ability to effectively process diverse product metadata, often resulting in hallucinations. In contrast, LLMs excel at generating coherent summaries that integrate both reviews and product specifications, earning consistent preference from human evaluators \cite{geval}.

%\textbf{Task-Agnostic Models:} Pre-trained models like {\tt BART-large} \cite{bart}, {\tt T5-large} \cite{t5}, and {\tt PEGASUS-large} \cite{pega} are constrained by their limited context windows ({\tt BART-large}: $1024$, {\tt T5:} $512-1024$, {\tt PEGASUS-large:} 4,096 tokens). These limitations lead to truncation and loss of critical product metadata, which often spans hundreds of words. Furthermore, these models lack the open-ended prompting capabilities and analytical reasoning required for effective M-OS. In contrast, LLMs' autoregressive nature and significantly larger context windows enable them to generate comprehensive summaries that maintain coherence while integrating diverse information sources, combining technical specifications with user experiences in a natural, user-friendly tone.

\textbf{Task-Agnostic Models:} Pre-trained models like {\tt BART-large} \cite{bart}, {\tt T5-large} \cite{t5}, and {\tt PEGASUS-large} \cite{pega} have limited context windows ({\tt BART-large}: 1024, {\tt T5:} 512-1024, {\tt PEGASUS-large:} 4,096 tokens), leading to truncation of critical product metadata. Unlike these models, LLMs' larger context windows and autoregressive nature enable comprehensive summaries that coherently integrate technical specifications with user experience.

\textbf{LLMs:} We evaluated models in a \textit{zero-shot} setting, as few-shot prompting requires significant human effort and is sensitive to example selection \cite{wan-etal-2023-better}. The complete list of models is provided in \textbf{(Appendix \ref{llms_list})}.

\begin{table*}[t]
    \centering
    \resizebox{2\columnwidth}{!}{%
    \begin{tabular}{clcccccccccccccc}
    \toprule
           &  \textbf{Evaluator LLM} & \multicolumn{2}{c}{\textbf{FL} $\uparrow$}  & \multicolumn{2}{c}{\textbf{CO} $\uparrow$} & \multicolumn{2}{c}{\textbf{FA} $\uparrow$} & \multicolumn{2}{c}{\textbf{RE} $\uparrow$} & \multicolumn{2}{c}{\textbf{AC} $\uparrow$} & \multicolumn{2}{c}{\textbf{SC} $\uparrow$} & \multicolumn{2}{c}{\textbf{SP} $\uparrow$}\\
         \cmidrule(lr){3-4} \cmidrule(lr){5-6} \cmidrule(lr){7-8} \cmidrule(lr){9-10} \cmidrule(lr){11-12} \cmidrule(lr){13-14} \cmidrule(lr){15-16}
         & & $\rho$ & $\tau$ & $\rho$ & $\tau$ & $\rho$ & $\tau$ & $\rho$ & $\tau$ & $\rho$ & $\tau$ & $\rho$ & $\tau$ & $\rho$ & $\tau$ \\  
    \midrule
    \multirow{10}{*}{\rotatebox[origin=c]{90}{\textbf{M-OS-DATA}}} 
       & OP-Llama-3.1-8B-Instruct & $0.59$ & $0.52$ & $\underline{0.61}$ & $0.41$ & $0.61$ & $0.46$ & $0.57$ & $\underline{0.44}$ & $0.58$ & $0.41$ & $\underline{0.68}$ & $\underline{0.59}$ & $0.64$ & $0.47$ \\
       & SPECTRA-Llama-3.1-8B-Instruct & $0.62$ & $0.50$ & $0.59$ & $0.47$ & $0.60$ & $0.42$ & $0.60$ & $0.46$ & $0.60$ & $0.42$ & $0.69$ & $0.57$ & $0.63$ & $0.49$ \\
       & OP-Mistral-7B-Instruct-v0.2 & $0.58$ & $0.50$ & $0.67$ & $0.50$ & $0.61$ & $0.46$ & $0.68$ & $\mathbf{0.54}$ & $0.58$ & $0.39$ & $0.67$ & $0.59$ & $0.60$ & $0.43$ \\
       & SPECTRA-Mistral-7B-Instruct-v0.2 & $0.61$ & $0.46$ & $\mathbf{0.68}$ & $0.50$ & $\mathbf{0.77}^*$ & $\mathbf{0.63}^*$ & $0.60$ & $0.43$ & $\mathbf{0.68}$ & $\mathbf{0.57}$ & $0.67$ & $\underline{0.55}$ & $0.54$ & $\underline{0.67}$ \\
       & OP-Mistral-7B-Instruct-v0.3 & $\underline{0.37}$ & $\underline{0.29}$ & $0.60$ & $0.51$ & $0.68$ & $0.57$ & $0.52$ & $0.41$ & $0.59$ & $0.44$ & $0.59$ & $0.47$ & $\underline{0.50}$ & $0.49$ \\
       & SPECTRA-Mistral-7B-Instruct-v0.3 & $0.40$ & $0.30$ & $0.60$ & $0.50$ & $0.60$ & $0.43$ & $0.61$ & $0.48$ & $0.67$ & $0.50$ & $0.67$ & $0.55$ & $0.54$ & $\underline{0.67}$ \\
       & OP-Llama-3.1-70B-Instruct & $0.68$ & $0.50$ & $0.67$ & $0.50$ & $0.61$ & $0.46$ & $0.68$ & $\mathbf{0.54}$ & $0.58$ & $0.39$ & $0.67$ & $0.59$ & $0.60$ & $0.43$ \\
       & SPECTRA-Llama-3.1-70B-Instruct & $\mathbf{0.77}^*$ & $\mathbf{0.61}^*$ & $\mathbf{0.68}$ & $0.48$ & $\mathbf{0.72}$ & $\mathbf{0.61}$ & $\mathbf{0.67}$ & $0.57$ & $\mathbf{0.71}$ & $0.52$ & $0.59$ & $0.49$ & $0.61$ & $\mathbf{0.82}$ \\
       & OP-GPT 4o & $0.63$ & $0.50$ & $0.62$ & $\mathbf{0.55}$ & $0.68$ & $0.57$ & $0.68$ & $0.57$ & $0.69$ & $0.54$ & $0.67$ & $0.55$ & $\mathbf{0.67}$ & $0.48$ \\
       & SPECTRA-GPT 4o & $0.70$ & $0.56$ & $0.68$ & $0.50$ & $\mathbf{0.77}^*$ & $\mathbf{0.63}^*$ & $\mathbf{0.73}^*$ & $\mathbf{0.65}^*$ & $0.73$ & $0.54$ & $0.67$ & $0.46$ & $0.65$ & $0.57$ \\
    \bottomrule
    \end{tabular}
    }
    \caption{Summary-level \textit{Spearman} ($\rho$) and \textit{Kendall Tau} ($\tau$) correlations between LLM evaluator scores and human judgments across $7$ evaluation dimensions for the \textsc{M-OS-DATA} dataset, comparing \textsc{Op-Prompts} and \textsc{Spectra-Prompts} approaches. \textls[-60]{\textsc{FL}} (Fluency), \textls[-60]{\textsc{CO}} (Coherence), \textls[-60]{\textsc{FA}} (Faithfulness), \textls[-30]{\textsc{RE}} (Relevance), \textls[-30]{\textsc{AC}} (Aspect Coverage), \textls[-30]{\textsc{SC}} (Sentiment Consistency), \textls[-30]{\textsc{SP}} (Specificity). Best performing values are boldfaced, and second best are underlined. $*$ represents significant performance (p-value $< 0.05$).}
 \label{tab:M-OS_evaluator_dep}
\end{table*}

\subsection{M-OS-EVAL (Summary evaluation)}\label{M-OS_eval}
\textbf{Baselines:} Traditional metrics like ROUGE (1,2,L) \cite{lin-2004-rouge}, BERTSCORE \cite{bertscore}, and BARTSCORE \cite{bartscore} were omitted due to weak correlation with human judgments and limited evaluation capabilities \cite{shen2023opinsummeval}. Recently, LLMs have been used as reference-free evaluators for NLG outputs \cite{fu2023gpt,geval}. We employed $4$ open-source LLMs and \texttt{GPT-4o} (closed-source) as baselines to assess M-OS across $7$ dimensions (Refer \textbf{Appendix \ref{eval_imp_detail}} for implementation details).
%\textbf{Baselines:} We omitted traditional metrics like ROUGE \cite{lin-2004-rouge}, BERTSCORE \cite{bertscore}, and BARTSCORE \cite{bartscore} due to poor human correlation and limited evaluation capabilities \cite{shen2023opinsummeval}. Instead, following recent work using LLMs as reference-free evaluators \cite{fu2023gpt,geval}, we employed $4$ open-source LLMs and \texttt{GPT-4o} to assess summaries across $7$ dimensions. (Refer \textbf{Appendix \ref{eval_imp_detail}} for implementation details.)
\begin{table}[H]
    \centering
    \resizebox{1\columnwidth}{!}{%
    \begin{tabular}{lcccccccc}
    \toprule
         \textbf{Model} & \textbf{FL} $\uparrow$ & \textbf{CO} $\uparrow$ & \textbf{AC} $\uparrow$ & \textbf{FA} $\uparrow$ & \textbf{RL} $\uparrow$ & \textbf{SC} $\uparrow$ & \textbf{SP} $\uparrow$ & \textbf{AVG} $\uparrow$\\
    \midrule
        {\tt Mistral-7B-Instruct-v0.3}    & \underline{4.95} & 4.15 & 4.0    & 4.1  & 4.0    & 4.11 & 3.56 & 4.124 \\
        {\tt Meta-Llama-3.1-8B-Instruct}  & \textbf{4.96} & 4.07 & 3.83 & 4.01 & 3.92 & 4.02 & 3.00    & 3.973 \\
        {\tt Mistral-7B-Instruct-v0.2}    & 4.93 & 4.1  & 3.96 & 4.08 & 3.97 & 3.97 & 3.45 & 4.066 \\
        {\tt gemma-7b-it}                 & 4.66 & 3.87 & 3.7  & 4.03 & 3.86 & 3.72 & 2.94 & 3.826 \\
        {\tt vicuna-7b-v1.5}              & 4.02 & 3.33 & 3.46 & 3.86 & 3.63 & 3.24 & 2.7  & 3.463 \\
        {\tt zephyr-7b-beta}              & 4.93 & 4.05 & 3.88 & 4.1  & 3.96 & 3.87 & 3.2  & 3.999 \\
        {\tt GPT 4o}                      & \underline{4.95} & 4.22 & \underline{4.03} & \textbf{4.15} & 4.02 & 4.0    & 3.81 & \underline{4.169} \\
        {\tt Gemma-2-9b-it}               & \underline{4.95} & 4.19 & 3.78 & 4.04 & 3.97 & 4.0    & 3.18 & 4.016 \\
        {\tt Mistral-Small-Instruct-2409} & \underline{4.95} & 4.18 & 3.88 & 4.01 & 3.99 & 3.92 & 3.44 & 4.053 \\
        {\tt Mixtral-8x7B-Instruct-v0.1}  & 4.91 & 4.11 & 3.75 & 3.98 & 3.92 & 3.82 & 3.07 & 3.937 \\
        {\tt Qwen2.5-7B-Instruct}    & \underline{4.95} & 4.21 & 4.01 & 4.03 & 4    & \underline{4.05} & 3.51 & 4.109 \\
        {\tt Qwen2.5-32B-Instruct}   & \underline{4.95} & \underline{4.26} & \underline{4.08} & 4.08 & \underline{4.04} & \underline{4.05} & 3.58 & 4.149 \\
        {\tt Qwen2.5-72B-Instruct}   & \textbf{4.98} & \underline{4.26} & \textbf{4.1}  & \underline{4.08} & \textbf{4.1}  & 4.0    & \textbf{3.78} & \textbf{4.186} \\
        {\tt Meta-Llama-3.1-70B-Instruct} & \underline{4.95} & 4.2  & 3.8  & 3.96 & 3.91 & 4.0    & 2.98 & 3.971 \\
    \bottomrule
    \end{tabular}
    }
    \caption{Model-wise averaged annotator ratings of \textsc{M-OS} along $7$ dimensions \textls[-60]{\textsc{\fl}}, \textls[-60]{\textsc{\co}}, \textls[-30]{\textsc{\rel}}, \textls[-60]{\textsc{\fa}}, \textls[-30]{\textsc{\asp}}, \textls[-30]{\textsc{\sent}}, \textls[-30]{\textsc{\spec}}. Best scores are in bold, second-best are underlined}. 
    \label{tab:M-OS_human}
\end{table}
\section{Results and Analysis} \label{results_analysis}
%We present our experimental results across two key aspects: (1) LLMs performance in generating summaries (M-OS-GEN) and (\textit{2}) LLMs as evaluators for these summaries (M-OS-EVAL). 
We analyze two aspects: (1) LLMs' summary generation performance (M-OS-GEN) and (\textit{2}) LLMs as summary evaluators (M-OS-EVAL).

\subsection{Model Performance for M-OS Generation}
%Table \ref{tab:M-OS_human} summarizes M-OS-GEN model evaluations, showing average annotator ratings across $7$ dimensions for $14$ models \textbf{(Figure \ref{fig:human_ann})}, demonstrating how different model characteristics influence M-OS generation performance.
Table \ref{tab:M-OS_human} summarizes M-OS-GEN model evaluations, showing average annotator ratings across $7$ dimensions for $14$ models, demonstrating how model size and architectural differences influence M-OS performance.
\paragraph{Overall Performance Analysis}
Among all models, \texttt{Qwen2.5-72B-Instruct} achieves the highest overall rating $(4.186)$, followed by \texttt{GPT-4o} $(4.169)$ and \texttt{Qwen2.5-32B-Instruct} $(4.149)$. Across dimensions, models show consistent excellence in \textls[-60]{\textsc{\fl}} but varied performance in \textls[-60]{\textsc{\spec}}, indicating that while models can generate grammatically correct summaries, they differ in their ability to provide detailed, precise product information.

\paragraph{Model Size Impact}
We observe a clear correlation between model size and performance. Larger models demonstrate superior performance in generating coherent and comprehensive summaries. \texttt{Qwen2.5-72B-Instruct} and \texttt{Meta-Llama-3.1-70B-Instruct} leverage their extensive parameters to capture nuanced relationships in product metadata. In contrast, \texttt{vicuna-7b-v1.5} struggles particularly with \textls[-60]{\textsc{\co}} and \textls[-60]{\textsc{\spec}}, especially for products with extensive specifications spanning hundreds of words. Similarly, \texttt{gemma-7b-it} and \texttt{gemma-2-9b-it} fall short in \textls[-60]{\textsc{\asp}} and \textls[-60]{\textsc{\fa}} compared to larger models.

\paragraph{Open-Source vs. Proprietary Models}
Open-source models have shown remarkable progress, with \texttt{Mistral-7B-Instruct-v0.3} achieving competitive performance against \texttt{GPT-4o}. While \texttt{GPT-4o} maintains slight advantages in \textls[-60]{\textsc{\co}} and \textls[-60]{\textsc{\fa}}, the diminishing gap demonstrates the viability of open-source alternatives for resource-constrained settings. Notably, \texttt{Qwen2.5-72B-Instruct}'s superior performance over \texttt{GPT-4o} challenges the conventional assumption about closed-source model superiority.
\begin{table*}[t]
    \centering
    \resizebox{2\columnwidth}{!}{%
    \begin{tabular}{clcccccccccccccc}
    \toprule
           &\textbf{Evaluator LLM} & \multicolumn{2}{c}{\textbf{FL} $\uparrow$}  & \multicolumn{2}{c}{\textbf{CO} $\uparrow$} & \multicolumn{2}{c}{\textbf{FA} $\uparrow$} & \multicolumn{2}{c}{\textbf{RE} $\uparrow$} & \multicolumn{2}{c}{\textbf{AC} $\uparrow$} & \multicolumn{2}{c}{\textbf{SC} $\uparrow$} & \multicolumn{2}{c}{\textbf{SP} $\uparrow$}\\
         \cmidrule(lr){3-4} \cmidrule(lr){5-6} \cmidrule(lr){7-8} \cmidrule(lr){9-10} \cmidrule(lr){11-12} \cmidrule(lr){13-14} \cmidrule(lr){15-16}
         & & $\rho$ & $\tau$ & $\rho$ & $\tau$ & $\rho$ & $\tau$ & $\rho$ & $\tau$ & $\rho$ & $\tau$ & $\rho$ & $\tau$ & $\rho$ & $\tau$ \\  
    \midrule
    \multirow{10}{*}{\rotatebox[origin=c]{90}{\textbf{M-OS-DATA}}} 
       & OP-I--Llama-3.1-8B-Instruct & $0.57$ & $0.49$ & $\underline{0.60}$ & $0.50$ & $0.60$ & $\underline{0.43}$ & $0.61$ & $0.48$ & $\mathbf{0.67}$ & $0.50$ & $\underline{0.67}$ & $0.55$ & $0.54$ & $\underline{0.67}$ \\
       & OMNI-Llama-3.1-8B-Instruct & $0.62$ & $0.50$ & $0.59$ & $0.47$ & $0.60$ & $0.42$ & $0.60$ & $0.46$ & $0.60$ & $0.42$ & $\mathbf{0.69}$ & $0.57$ & $0.63$ & $0.49$ \\
       & OP-I--Mistral-7B-Instruct-v0.2 & $0.62$ & $0.42$ & $\mathbf{0.67}$ & $0.50$ & $0.63$ & $0.48$ & $0.68$ & $\mathbf{0.54}$ & $0.58$ & $0.50$ & $0.67$ & $0.59$ & $0.62$ & $0.43$ \\
       & OMNI-Mistral-7B-Instruct-v0.2 & $\mathbf{0.67}$ & $0.50$ & $\mathbf{0.67}$ & $0.55$ & $\mathbf{0.68}$ & $0.57$ & $0.68$ & $0.57$ & $0.69$ & $\mathbf{0.54}$ & $\mathbf{0.67}$ & $\mathbf{0.55}$ & $\mathbf{0.67}$ & $\mathbf{0.48}$ \\
       & OP-I--Mistral-7B-Instruct-v0.3 & $0.59$ & $\mathbf{0.52}$ & $\underline{0.61}$ & $0.41$ & $0.61$ & $\underline{0.46}$ & $0.57$ & $\underline{0.44}$ & $0.58$ & $0.50$ & $0.68$ & $0.59$ & $\underline{0.64}$ & $0.47$ \\
       & OMNI-Mistral-7B-Instruct-v0.3 & $0.58$ & $0.50$ & $0.67$ & $0.50$ & $\underline{0.61}$ & $\underline{0.46}$ & $\mathbf{0.68}$ & $\mathbf{0.54}$ & $0.67$ & $0.55$ & $0.60$ & $0.43$ & $\underline{0.60}$ & $0.43$ \\
       & OP-I--Llama-3.1-70B-Instruct & $0.68$ & $0.50$ & $\mathbf{0.67}$ & $0.50$ & $0.73$ & $\mathbf{0.59}$ & $0.78$ & $\mathbf{0.64}$ & $\mathbf{0.67}$ & $\mathbf{0.57}$ & $0.67$ & $0.46$ & $\mathbf{0.65}$ & $0.57$ \\
       & OMNI-Llama-3.1-70B-Instruct & $\mathbf{0.70}$ & $\mathbf{0.56}$ & $0.68$ & $\mathbf{0.55}$ & $\mathbf{0.77}^*$ & $\mathbf{0.63}^*$ & $0.73$ & $\mathbf{0.65}$ & $\mathbf{0.71}$ & $0.54$ & $0.67$ & $0.46$ & $0.61$ & $\mathbf{0.82}$ \\
       & OP-I--GPT 4o & $0.69$ & $0.53$ & $0.67$ & $0.61$ & $0.68$ & $0.57$ & $0.79$ & $0.56$ & $0.71$ & $0.54$ & $0.67$ & $0.55$ & $0.67$ & $0.48$ \\
       & OMNI-GPT 4o & $\mathbf{0.76}^*$ & $\mathbf{0.59}^*$ & $\mathbf{0.72}$ & $\mathbf{0.61}$ & $\mathbf{0.77}^*$ & $\mathbf{0.63}^*$ & $\mathbf{0.82}^*$ & $\mathbf{0.65}^*$ & $0.74$ & $\mathbf{0.62}$ & $0.68$ & $\mathbf{0.46}$ & $0.66$ & $\mathbf{0.46}$ \\
    \bottomrule
    \end{tabular}
    }
    \caption{Summary-level \textit{Spearman} ($\rho$) and \textit{Kendall Tau} ($\tau$) correlations between LLM evaluator scores and human judgments across $7$ evaluation dimensions for the \textsc{M-OS-DATA} dataset, comparing \textsc{Op-I-Prompt} and \textsc{Omni-Prompt} approaches. \textls[-60]{\textsc{FL}} (Fluency), \textls[-60]{\textsc{CO}} (Coherence), \textls[-60]{\textsc{FA}} (Faithfulness), \textls[-30]{\textsc{RE}} (Relevance), \textls[-30]{\textsc{AC}} (Aspect Coverage), \textls[-30]{\textsc{SC}} (Sentiment Consistency), \textls[-30]{\textsc{SP}} (Specificity). Best performing values are boldfaced, and second best are underlined. $*$ represents significant performance (p-value $< 0.05$).}
    \label{tab:_M-OS_evaluator_inp}
\end{table*}

\paragraph{The Qwen Family Performance}
The Qwen family, particularly \texttt{Qwen2.5-72B-Instruct} and \texttt{Qwen2.5-32B-Instruct}, excels across all dimensions. Interestingly, \texttt{Qwen2.5-7B-Instruct} shows strong performance despite its smaller size, particularly in \textls[-60]{\textsc{\fa}} and \textls[-60]{\textsc{\rel}}, indicating that careful tuning can partially compensate for model size limitations.

\subsection{LLMs as M-OS Evaluators}
Table \ref{tab:M-OS_evaluator_dep} and \ref{tab:_M-OS_evaluator_inp} demonstrate the evaluation capabilities of LLMs.
\noindent
\paragraph{\textsc{\textbf{Op-Prompts}} vs. \textsc{\textbf{Spectra-Prompts}}.}  The results, summarized in Table \ref{tab:M-OS_evaluator_dep}, present summary-level correlations for various models evaluated using %\textsc{\textbf{Op-Prompts}} and \textsc{\textbf{Spectra-Prompts}}
metric-dependent prompts on the \textsc{M-OS-DATA} dataset. Overall, \textls[-30]{\textsc{Spectra-GPT-4o}} achieves the best performance with an average Spearman correlation of $0.70$ across all dimensions, followed by \textls[-30]{\textsc{Spectra-Llama-3.1-70B-Instruct}} ($0.68$) and \textls[-30]{\textsc{Spectra-Mistral-7B-Instruct-v0.2}} ($0.65$). Notably, \textsc{\textbf{Spectra-Prompts}} \textit{consistently outperform} \textsc{\textbf{Op-Prompts}} \textit{across all LLMs acting as evaluators}, demonstrating the effectiveness of dimension-specific prompting strategies.

\paragraph{\textsc{\textbf{Op-I-Prompt}} vs. \textsc{\textbf{Omni-Prompt}}.} 
The results, summarized in Table \ref{tab:_M-OS_evaluator_inp}, present summary-level correlations for various models evaluated using metric-independent prompts on the \textsc{M-OS-DATA} dataset. \textls[-30]\textsc{Omni-Prompt} with \textls[-30]{\textsc{GPT-4o}} as the backbone achieves the strongest performance with an average Spearman correlation of $0.74$ across all dimensions, followed by \textls[-30]{\textsc{Omni-Llama-3.1-70B-Instruct}} ($0.70$) and \textls[-30]{\textsc{Omni-Mistral-7B-Instruct-v0.2}} ($0.68$). Notably, \textls[-30]\textsc{\textbf{Omni-Prompt}} \textit{consistently outperforms} \textls[-30]\textsc{\textbf{Op-I-Prompt}} \textit{across all LLMs acting as evaluators}, demonstrating the effectiveness of our metric-independent prompting strategy.

\noindent \paragraph{Dimension-wise Analysis:} For metric-dependent evaluation, \textls[-30]{\textsc{Spectra-GPT-4o}} shows significant improvements in \textls[-60]{\fa} and \textls[-30]{\rel}, while for metric-independent evaluation, \textls[-30]{\textsc{Omni-GPT-4o}} achieves significant gains in \textls[-60]{\fl}, \textls[-60]{\fa}, and \textls[-30]{\rel}. Both prompting strategies demonstrate strong performance in \textls[-60]{\fa} and \textls[-30]{\rel}. Additionally, \textls[-30]{\asp} evaluation shows competitive performance across model sizes, highlighting the effectiveness of structured prompting.

\paragraph{Closed vs. Open-Source Models:} While proprietary models like \textls[-30]{\texttt{GPT-4o}} show stronger alignment with human judgments, open-source alternatives like \textls[-30]{\texttt{Llama-3.1-70B-Instruct}} and \textls[-30]{\texttt{Mistral-7B-Instruct-v0.2}} demonstrate competitive performance, indicating their viability for resource-constrained environments. However, smaller models like \textls[-30]{\texttt{Llama-3.1-8B-Instruct}} underestimate scores while \textls[-30]{\texttt{Mistral-7B-Instruct-v0.3}} inflates them, reducing human judgment correlations.

\paragraph{Comparative Analysis of Prompting Strategies.} Our detailed analysis of model responses between our proposed approaches (\textsc{Spectra-Prompts} and \textls[-30]\textsc{Omni-Prompt}) and baseline approaches (\textsc{Op-Prompts} and \textsc{Op-I-Prompt})\footnote{Model responses are provided in our supplementary materials \url {https://anonymous.4open.science/r/M-OS-D3C3/}.} reveals two key findings across both metric-dependent and metric-independent evaluations: (\textit{1}) baseline prompts show score inflation compared to our approaches, and (\textit{2}) our structured prompting enforces rigorous evaluation through identifying summary elements, conducting systematic analysis, and determining scores using defined percentage ranges, while baseline approaches' less structured methodology leads to score overestimation.

\section{User Study: M-OS Effectiveness} \label{study}
We conducted a user study with $300$ participants (aged $18-50$) to evaluate the quality of summaries generated by the top-performing model, \texttt{Qwen2.5-72B-Instruct} (Table~\ref{tab:M-OS_human}). Participants compared $4$ pairs of summaries: \textsc{M-OS} vs. traditional opinion-summary method. To eliminate bias, the summaries were neutrally labeled as \texttt{"Summary 1"} and \texttt{"Summary 2"}. On average, participants preferred \textsc{M-OS} summaries \textbf{87\% of the time} across five evaluation criteria, highlighting both the theoretical soundness and practical effectiveness of our method. Detailed survey information is provided in \textbf{Appendix~\ref{study_detail}}.

\section{Conclusion and Future Work}
%In this work, we extended multi-source opinion summarization framework (M-OS) by leveraging LLMs to generate comprehensive summaries that integrate diverse product metadata with customer reviews. The framework demonstrates superior performance through: (\textit{1}) \textsc{M-OS-DATA}, a novel proprietary dataset comprising $25,000$ products with rich metadata, (\textit{2}) \textsc{M-OS-EVAL}, a benchmark dataset with $4,900$ summary annotations across $7$ key dimensions, and (\textit{3}) custom-tailored prompts for \textls[-60]{\textsc{M-OS-GEN}} and two novel evaluation prompts:\textsc{Omni-Prompt} and \textsc{Spectra-Prompts}. Our experiments demonstrate that LLM-generated M-OS achieves strong alignment with human judgment, with an average Spearman correlation of $\textbf{0.74}$ across evaluation dimensions. In future work, we plan to explore LLMs for large-scale comparative opinion summarization and investigate novel approaches for handling the complete review corpus.

In this work, we extend multi-source opinion summarization (M-OS) by leveraging LLMs to generate comprehensive summaries integrating product metadata with customer reviews. Our framework introduces: (\textit{1}) \textsc{M-OS-DATA}, a proprietary dataset of $25,000$ products with rich metadata, (\textit{2}) \textsc{M-OS-EVAL}, a benchmark dataset of 4,900 summary annotations across $7$ dimensions, and (\textit{3}) custom prompts for \textls[-60]{\textsc{M-OS-GEN}} with two novel evaluation approaches: \textsc{Omni-Prompt} and \textsc{Spectra-Prompts}. Experiments show \textsc{M-OS} achieves strong alignment with human judgment, demonstrating a $\mathbf{0.74}$ average Spearman correlation across dimensions. 
While effectively combining product specifications with reviews, our approach needs expansion to handle temporal patterns and multi-modal content in modern e-commerce. Both evaluation prompts require further testing across languages and cultures, as opinion expression varies globally. Future work will explore LLMs for large-scale summarization and processing complete review corpora.

\section*{Limitations}

\begin{enumerate}
    \item Our study focused on \textsc{GPT-4o}, the only proprietary model used in our experiments. We did not include \textsc{Claude-Sonnet 3.5} \cite{claude35sonnet} due to budget limitations.

  \item Creating the \textsc{M-OS-EVAL} dataset advances multi-source opinion summarization research, with future work exploring model fine-tuning opportunities on this dataset.
    
    \item Our \textsc{Omni-Prompt} effectively evaluates multiple dimensions of multi-source opinion summaries, while \textsc{Spectra-Prompts} excel at dimension-specific evaluation. However, adapting these prompts beyond opinion summarization to other \textsc{NLP} tasks would require domain-specific modifications to the evaluation criteria.

   \item The current \textsc{M-OS-EVAL} benchmark dataset, while comprehensive in its evaluation dimensions, is based on 10 reviews per product. While this offers meaningful insights, expanding the dataset to encompass a larger and more diverse set of reviews would better capture the variety and complexity of real-world e-commerce scenarios.

    %\item Our framework, while effective for products with structured metadata and moderate review volumes, requires further investigation for scalability to real-world e-commerce platforms where products may have thousands of reviews and incomplete metadata.

\end{enumerate}

\section*{Ethical Considerations}
We prioritized responsible development and evaluation throughout our research. The evaluation process involved $3$ experienced raters (Master's, Pre-Doctoral, Doctoral) aged $24-32$, all with relevant publications or active research in opinion summarization. We ensured ethical data practices by obtaining \textsc{M-OS-DATA} through formal collaboration with an e-commerce company, following strict quality controls and privacy protocols.

To maintain evaluation integrity, we: (\textit{1}) provided raters with detailed annotation guidelines and appropriate compensation, (\textit{2}) kept model identities undisclosed during evaluation, and (\textit{3}) established clear metrics and scoring criteria across $7$ dimensions. Our \textsc{M-OS-PROMPTS} framework, while designed to assist researchers and developers in assessing NLG-generated summaries, has certain limitations. The evaluation prompts may occasionally produce hallucinations, particularly for complex cases, and the LLM-based approach may exhibit inherent biases.

We advise practitioners to: (\textit{1}) validate prompt reliability before real-world deployment, (\textit{2}) verify prompt appropriateness for specific applications, and (\textit{3}) consider potential limitations when interpreting results. This transparency about limitations and guidelines for responsible use ensures ethical application of our framework in research and practical settings.

\bibliography{anthology,custom}

\appendix
\section{M-OS Metrics} \label{metric_definition}
The evaluation of multi-source opinion summaries was conducted across the following $7$ dimensions:
\begin{enumerate}
    \item {\tt \textbf{fluency}} \textbf{(FL)}- Fluency measures the quality of the summary in terms of grammar, spelling, punctuation, capitalization, word choice, and sentence structure. The summary should be easy to read, follow, and comprehend without any errors that hinder understanding. Annotators received specific guidelines on how to penalize
summaries based on fluency levels.
    \item {\tt \textbf{coherence}} \textbf{(CO)}- Coherence measures the collective quality of all sentences in the summary. The summary should be well-structured and well-organized. It should not just be a heap of related information, but should build from sentence to sentence into a coherent body of information about the product.
    \item {\tt \textbf{relevance}} \textbf{(RE)}- Relevance measures the selection of important information from the input, including product title, description, key features, specifications, reviews, and average rating. The summary should include only important and relevant information from the input. Summaries should not contain redundancies or excess information. Annotators were instructed to penalize summaries if they contained redundancies and excess/unimportant information.
    \item {\tt \textbf{faithfulness}} \textbf{(FA)}- Faithfulness measures the extent to which every piece of information mentioned in the summary is verifiable, supported, present, or can be reasonably inferred from the input. The input includes product title, description, key features, specifications, reviews, and average rating. Summaries should be penalized if they contain information that cannot be verified from the provided input or if they make broad generalizations that are not supported by the input data.
    \item {\tt \textbf{aspect coverage}} \textbf{(AC)}- Aspect Coverage measures how completely a summary captures the major features, characteristics, or attributes of a product that are prominently discussed in the original product information. Summaries should be penalized for missing any major aspects and rewarded for covering all important aspects thoroughly.
    \item {\tt \textbf{sentiment consistency}} \textbf{(SC)}- Sentiment Consistency measures how accurately the summary reflects the consensus sentiment of users for each aspect of the product as expressed in the reviews. The consensus sentiment (or majority sentiment) for an aspect is determined by the M-OSt common sentiment expressed by users, categorized as very positive, positive, neutral, negative, or very negative. Summaries should be penalized if they do not cover accurately the sentiment regarding any aspect within the summary.
    
    \item {\tt \textbf{specificity}} \textbf{(SP)}- Specificity measures the level of detail and precision in the information and opinions presented in the summary. A specific summary provides concrete facts, measurements, or detailed descriptions about the product's features, performance, and user experiences. It avoids vague or general statements and instead offers precise information that gives readers a clear and thorough understanding of the product's characteristics and performance. Summaries should be penalized for missing out details and should be awarded if they are specific.
    
\end{enumerate}

\section{Prompt Design Principle}\label{design_principle}
We define design principles for M-OS-PROMPTS as follows:

\subsection{M-OS-GEN-PROMPT Design Principle} \label{gen_design_principle}
The design of \textsc{M-OS-GEN-PROMPT} is based on the principle of clarity and balance. The intuition behind this approach is that explicitly defining each aspect of the task enables the generation of summaries that are both accurate and easy to understand. Our prompt design emphasizes:

\noindent(\textit{1}) {\tt \textbf{Information Balance:}} The prompt requires balanced integration of objective product data (specifications, features) with subjective feedback (reviews, ratings), ensuring no single attribute dominates unless warranted by the data.

\noindent(\textit{2}) {\tt \textbf{Structured Coverage:}} Each summary sentence must focus on a distinct product aspect with specific details, avoiding redundancy while maintaining comprehensive coverage.

\noindent(\textit{3}) {\tt \textbf{Accessibility:}} The generated summaries use clear, professional language while avoiding technical jargon, making them useful for quick decision-making in e-commerce contexts.

\subsection{M-OS-EVAL-PROMPTS Design Principle} \label{eval_design_principle}
The design of our evaluation prompts is grounded in the intuition that LLMs generate more robust responses when required to justify their evaluations. Our approach ensures that the response explicitly reiterates the evaluation metric, emphasizes both the strengths and shortcomings of the summary, and concludes with an evaluation score aligned with the criteria specified in the prompt. Our design emphasizes:

\noindent(\textit{1}) {\tt \textbf{Comprehensive Evaluation:}} The prompts are structured to assess summaries that integrate multiple sources of product information, considering both objective product specifications and subjective customer feedback in a unified evaluation framework.

\noindent(\textit{2}) {\tt \textbf{Structured Assessment Framework:}} Each evaluation follows a systematic approach with clear definition of the evaluation dimension, step-by-step analysis against provided criteria, quantified scoring with explicit justification, and standardized score reporting using <score></score> tags.

\noindent(\textit{3}) {\tt \textbf{Guided Scoring Mechanism:}}  We introduce precise percentage ranges for each score level, providing LLMs with clear benchmarks for evaluation. This prevents score inflation or deflation by giving LLMs concrete criteria for assessment. Additionally, we present evaluation criteria in a structured bullet-point format for each score level, as we observed this format leads to more consistent and accurate evaluations compared to paragraph-style descriptions.

\noindent(\textit{4}) {\tt \textbf{Adaptive Architecture:}} The prompts support different evaluation needs through OMNI-PROMPT's modular design for metric-independent evaluation and Spectra-Prompts' specialized criteria for dimension-specific assessment. This dual approach ensures both flexibility and precision in evaluating multi-source opinion summaries.

\section{User Study Details} \label{study_detail}
In this section we describe the detailed analysis of user study:
\subsection*{D.1 Study Design and Methodology}

We conducted a large-scale user study ($N = 300$) comparing \textsc{M-OS} (Multi-Source Opinion Summaries) with traditional opinion summaries. Each participant evaluated four pairs of summaries.

Participants provided responses to five evaluation questions, resulting in $6,000$ total preference judgments ($300$ participants $\times 4$ categories $\times 5$ questions).

\subsection*{D.2 Evaluation Questions}

\begin{itemize}
    \item \textbf{Information Comprehensiveness:} ``Which summary type (\textsc{M-OS} or Opinion Summary) provides a more complete understanding of both product specifications and customer experiences?''
    \item \textbf{Decision Confidence:} ``Which summary format gives you more confidence in understanding the product's actual capabilities and limitations?''
    \item \textbf{Specification Understanding:} ``Which summary better helps you understand both technical specifications and real-world performance?''
    \item \textbf{Research Efficiency:} ``Which summary would reduce your need to look up additional product information elsewhere?''
    \item \textbf{Purchase Decision Support:} ``Which summary format provides a better balance of technical details and user experiences to support your purchase decision?''
\end{itemize}

\subsection*{D.3 Statistical Analysis}

To validate the statistical significance of user preferences, we employed the chi-square goodness-of-fit test, following established practices in NLP user studies \citep{clark-etal-2011-better}. This test assessed whether the observed preference distribution significantly deviated from the null hypothesis of no preference (a 50-50 split).

\begin{figure}[t]
    \centering
    \includegraphics[width=1\columnwidth]{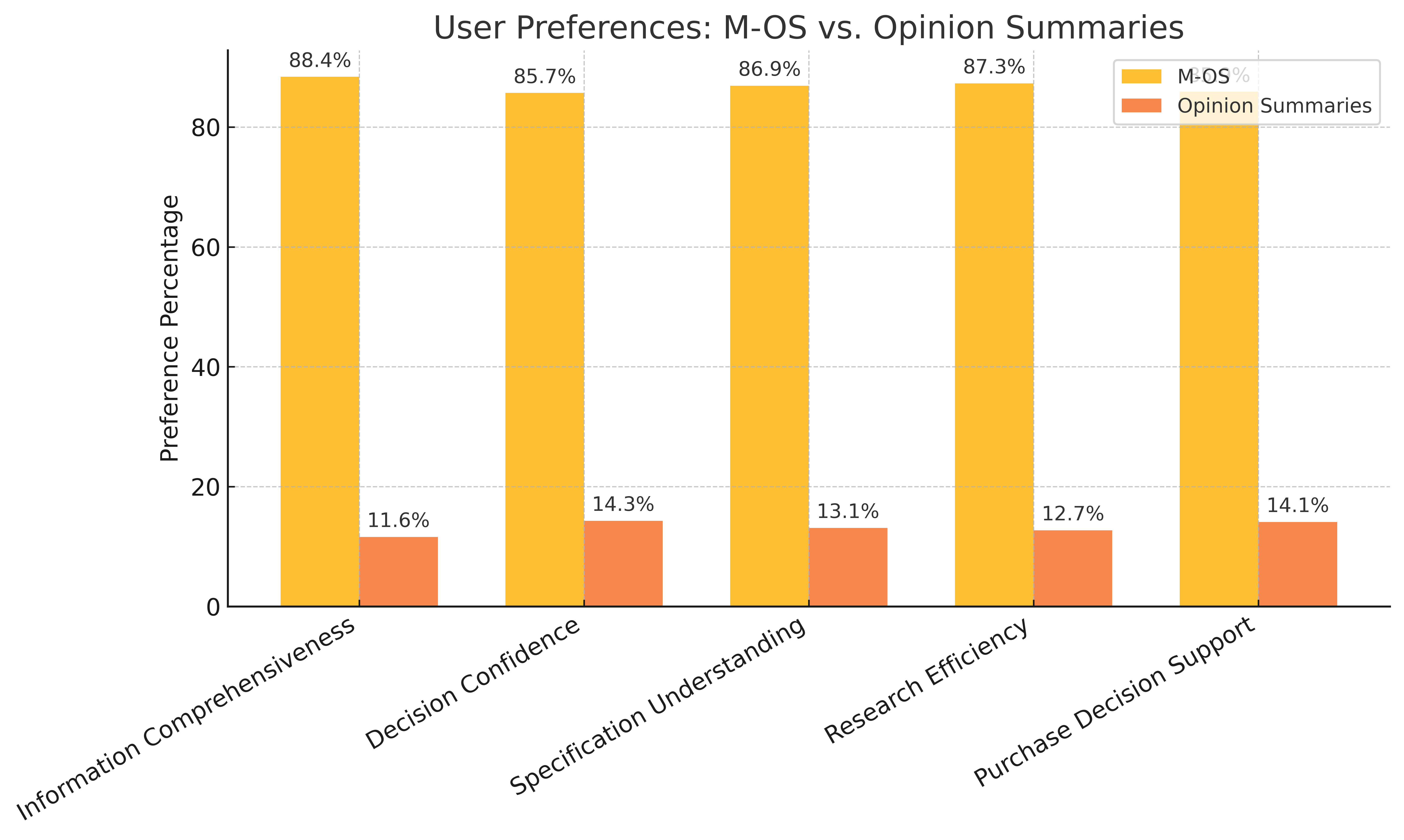}
    \caption{Preference analysis comparing Multi-Source Opinion Summaries (\textsc{M-OS}) versus traditional opinion summaries across product categories ($N=300$). Bars represent the mean preference percentage across five evaluation questions per category. Statistical significance: $\chi^2 = 3126.83$ ($df = 1$, $p < .001$).}
    \label{fig:preference_analysis}
\end{figure}

The chi-square test produced $\chi^2 = 3126.83$ ($df = 1$, $p < .001$), strongly rejecting the null hypothesis of no preference. The observed preference distribution across evaluation criteria was as follows:

\begin{itemize}
    \item \textbf{Information Comprehensiveness:} 88.4\% (\textsc{M-OS}) vs. 11.6\% (traditional)
    \item \textbf{Decision Confidence:} 85.7\% vs. 14.3\%
    \item \textbf{Specification Understanding:} 86.9\% vs. 13.1\%
    \item \textbf{Research Efficiency:} 87.3\% vs. 12.7\%
    \item \textbf{Purchase Decision Support:} 85.9\% vs. 14.1\%
\end{itemize}

Overall, participants expressed 5,196 preferences for \textsc{M-OS} (86.6\%) compared to 804 preferences for traditional opinion summaries (13.4\%), providing strong evidence of \textsc{M-OS}’s superiority.

To measure the strength of this effect, we calculated Cramer's $V = 0.72$, indicating a large effect size based on conventional behavioral research benchmarks \citep{cohen1988statistical}.

\section{LLMs Utilized}\label{llms_list}
In our experiments, we adopt a range of recent widely-used LLMs. For close-sourced LLM (accessible through APIs), we evaluate OpenAI’s {\gptfourstandard} \cite{openai}. For open-source LLMs, use the HuggingFace library \cite{hf} to access these models and experimented with {\tt Mistral-7B-Instruct-v0.2} \cite{m2}, {\tt Mistral-7B-Instruct-v0.3} \cite{m2}, {\tt gemma-7b-it} \cite{gemma}, {\tt vicuna-7b-v1.5} \cite{v}, {\tt zephyr-7b-beta} \cite{z}, {\tt Qwen2.5-7B-Instruct} \cite{q}, {\tt Meta-Llama-3.1-8B-Instruct} \cite{l3}, {\tt Gemma-2-9b-it} \cite{gemma}, {\tt Mistral-Nemo-Instruct-2407} \cite{m2}, {\tt Qwen2.5-14B-Instruct} \cite{q}, {\tt Mistral-Small-Instruct-2409} \cite{m2}, {\tt Mixtral-8x7B-Instruct-v0.1} \citet{mix}, {\tt Qwen2.5-32B-Instruct} \cite{q}, {\tt Meta-Llama-3.1-70B-Instruct}, {\tt Qwen2.5-72B-Instruct} \cite{q}.

\section{Implementation Details} \label{imp_detail}
All experiments were conducted on $8$ NVIDIA $A100-SXM4-80GB$ clusters, providing ample computational power for robust analyses.
\subsection{M-OS-GEN Implementation Details:} \label{gen_imp_detail}
For inference, we configured both closed-source and open-source LLMs. After extensive experimentation, we selected \verb|top_k=25, top_p=0.95| and \verb|temperature=0.2| to generate deterministic, coherent outputs that effectively capture the comprehensive fine-grained product details, ensuring consistent and reliable performance across all models. 

\subsection{M-OS-EVAL Implementation Details:} \label{eval_imp_detail}
To ensure robust evaluation and account for potential stochasticity in LLM outputs, we set $n=100$ , evaluating each summary $100$ times across both closed and open-source LLMs. A temperature of 0.0 was used to ensure deterministic outputs, aiming for consistent, high-quality results crucial for reproducibility and reliable evaluation.

\section{Dependent and Independent Prompts Visualization} \label{prompt_view}
\begin{figure*}[htp]
    \centering
    \includegraphics[width=2\columnwidth]{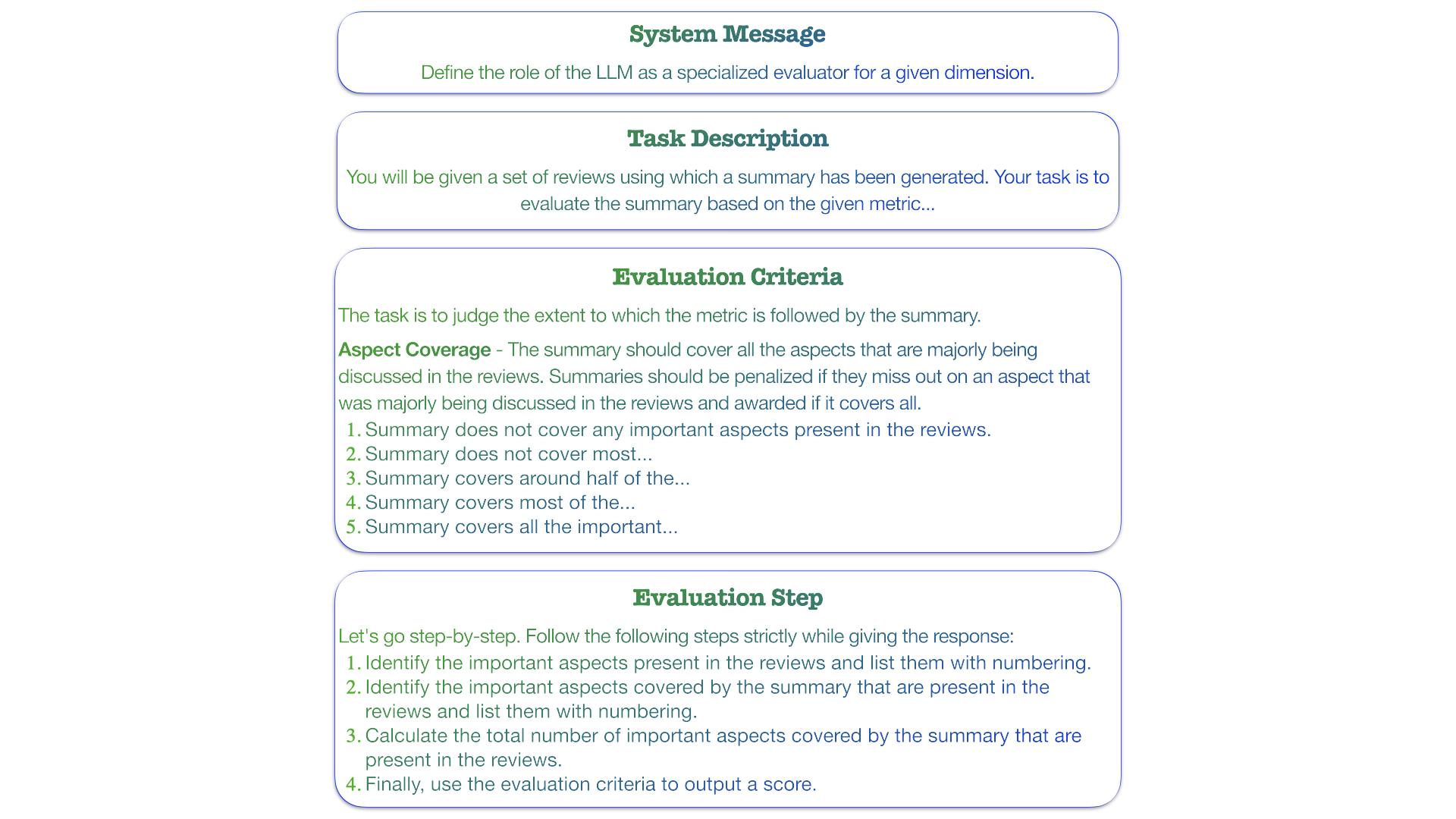}
    \caption{Structure of Dependent Prompts}
    \label{fig:d}
\end{figure*}

We provide a graphical overview of the dependent prompts in Figure \ref{fig:d} and independent prompts in Figure \ref{fig:i}, illustrating their structure, components, and evaluation criteria. The visualizations highlight the differences in prompt design, emphasizing the step-by-step evaluation process and how metrics are applied for assessing summaries.

\begin{figure*}[t]
    \centering
    \includegraphics[width=2\columnwidth]{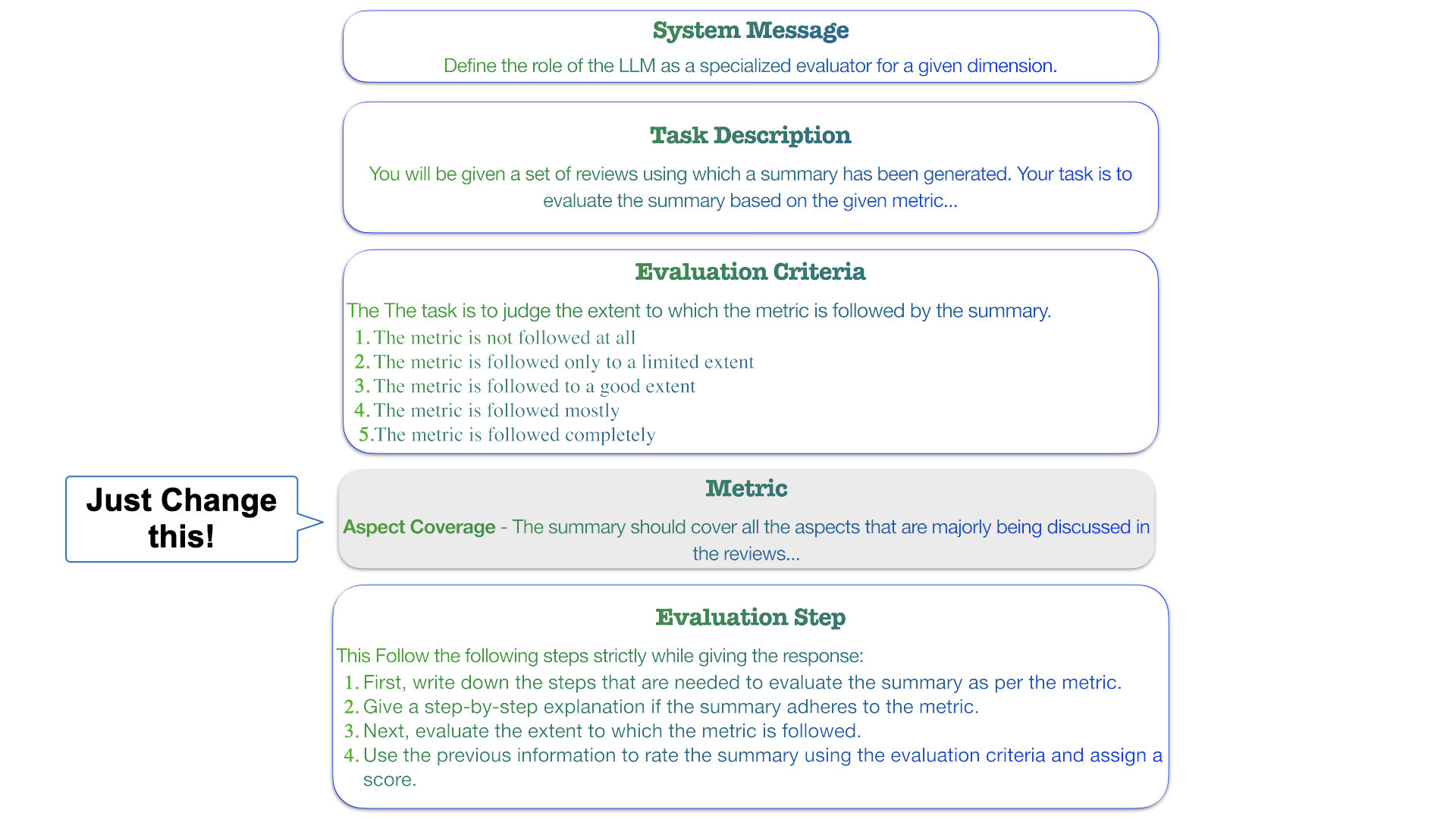}
    \caption{Structure of Inependent Prompts} 
    \label{fig: i}
\end{figure*}

\end{document}